\documentclass[letterpaper, 10 pt, conference]{ieeeconf}  
\usepackage{color}
\newcommand{\ie}{i.e.,\ }

\newcommand{\Reffig}[1]{Fig.~\ref{#1}}
\newcommand{\Refsec}[1]{Sec.~\ref{#1}}

\newcommand{\Refeq}[1]{Eq.~\ref{#1}}
\newcommand{\Reftab}[1]{Tab.~\ref{#1}}

\usepackage{graphics} 
\usepackage{epsfig} 
\usepackage{algorithm}
\usepackage{algpseudocode}
\usepackage{amsmath}
\usepackage{amssymb}
\usepackage{subfigure}
\usepackage{graphicx}
\usepackage{subfigure}
\usepackage{booktabs}
\usepackage{threeparttable}
\usepackage{multirow}
\usepackage{multicol}
\usepackage{makecell}
\usepackage{dcolumn}
\usepackage[T1]{fontenc}
\usepackage{diagbox}
\usepackage[normalem]{ulem}
\useunder{\uline}{\ul}{}

\IEEEoverridecommandlockouts                              

\title{\LARGE \bf
Convex Hull-based Algebraic Constraint for Visual Quadric SLAM
}

\author{Xiaolong Yu$^{1}$, Junqiao Zhao$^{*,1, 2}$, Shuangfu Song$^{3}$, Zhongyang Zhu$^{1}$, Zihan Yuan$^{1}$, Chen Ye$^{1}$, Tiantian Feng$^{3}$
\thanks{This work is supported by the National Key Research and Development Program of China (No. 2021YFB2501104). \emph{(Corresponding Author: Junqiao Zhao.)}}
\thanks{$^{1}$Xiaolong Yu, Junqiao Zhao, Zhongyang Zhu, Zihan Yuan and Chen Ye are with the School of Computer Science and Technology, 
Tongji University, Shanghai, China, and the MOE Key Lab of Embedded System and Service Computing, Tongji University, Shanghai, China 
{\tt\footnotesize (e-mail: 2230795@tongji.edu.cn; zhaojunqiao@tongji.edu.cn; 2233057@tongji.edu.cn; 2332062@tongji.edu.cn; yechen@tongji.edu.cn).}}
\thanks{$^{2}$Institute of Intelligent Vehicles, Tongji University, Shanghai, China}
\thanks{$^{3}$Shuangfu Song and Tiantian Feng are with the School of Surveying and Geo-Informatics, Tongji University, Shanghai, China
{\tt\footnotesize (e-mail: 1911202@tongji.edu.cn; fengtiantian@tongji.edu.cn).}}
}

\begin{document}

\maketitle
\thispagestyle{empty}
\pagestyle{empty}

\begin{abstract}
   Using Quadrics as the object representation has the benefits of both generality and closed-form projection derivation between image and world spaces. 
   Although numerous constraints have been proposed for dual quadric reconstruction, we found that many of them are imprecise and provide minimal improvements to localization.
   After scrutinizing the existing constraints, we introduce a concise yet more precise convex hull-based algebraic constraint for object landmarks, which is applied to object reconstruction, frontend pose estimation, and backend bundle adjustment.
   This constraint is designed to fully leverage precise semantic segmentation, effectively mitigating mismatches between complex-shaped object contours and dual quadrics. 
   Experiments on public datasets demonstrate that our approach is applicable to both monocular and RGB-D SLAM and achieves improved object mapping and localization than existing quadric SLAM methods. 
   The implementation of our method is available at https://github.com/tiev-tongji/convexhull-based-algebraic-constraint. 
\end{abstract}

\section{Introduction}
In recent years, with the rapid development of object detection and semantic segmentation, 
many object-based SLAM systems have been proposed \cite{nicholson2018quadricslam, yang2019cubeslam, wu2020eao, zins2022oa, wang2023qiso}. 
By mapping and localizing high-level object landmarks, the robustness of the SLAM system is improved because the image feature points are susceptible to environmental degradation or variations such as textureless regions and illumination changes. 
Additionally, object-based mapping enhances scene understanding and human-computer interaction capabilities.


QuadricSLAM \cite{nicholson2018quadricslam} first proposes using dual quadric as the object representation in visual SLAM due to its generality and rigorous projection properties between the image and world spaces \cite{hartley2003multiple}.
Subsequently, various constraints have been proposed for the reconstruction of dual quadrics, including the bounding box (bbox)-derived constraints \cite{nicholson2018quadricslam, ok2019robust, hosseinzadeh2019real, tian2021accurate, sunderhauf2017dual}, conic-derived constraint \cite{zins2022oa} and contour-derived constraints \cite{wang2023qiso}.
However, we found that most existing quadric SLAM methods focus on accurate semantic object reconstruction but fail to leverage these objects to improve localization. 
The reason is that existing constraints between dual quadrics and observations are insufficiently precise to improve pose estimation.

\begin{figure}[t]
   \centering
   \includegraphics[width=0.8\linewidth]{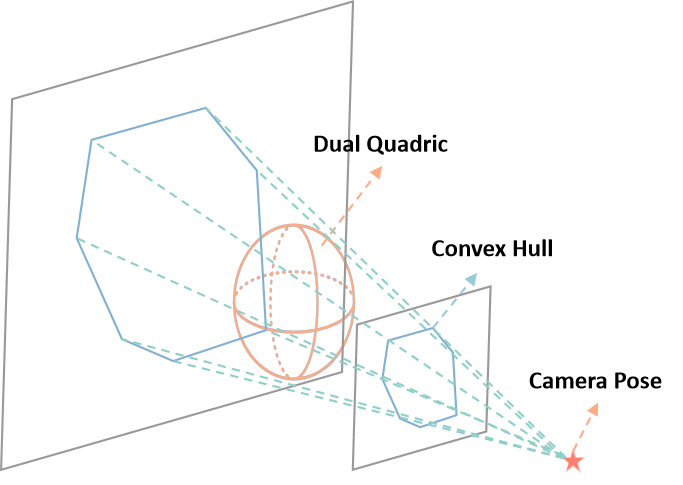}
   \caption{The Proposed Convex Hull-based Algebraic Constraint between Dual Quadric and Camera Pose.}
   \label{fig_ourconstraint}
   \vspace{-0.4cm}
\end{figure}

To address these issues, we evaluate existing constraints for quadric SLAM, and 
propose a convex hull-based algebraic constraint that fully leverages precise instance segmentation, as illustrated in \Reffig{fig_ourconstraint}. 
This approach refines rough bbox or conic-based object observation and mitigates mismatches between complex-shaped object contours and their corresponding dual quadrics, thereby enhancing multi-view consistency.

To effectively apply the proposed constraint for camera localization, we integrate it into both frontend pose estimation and backend Bundle Adjustment (BA) within a quadric SLAM system. 


Experimental results demonstrate that our proposed method significantly improves both localization accuracy and object mapping performance in monocular and RGB-D visual SLAM systems, outperforming existing quadric-based methods across multiple sequences.

The main contributions of this work are as follows:

\begin{itemize}

   \item A convex hull-based algebraic constraint is proposed to fully leverage accurate segmentation contours and mitigate the mismatches between complex-shaped objects and their dual quadrics. 
   
   \item A monocular/RGB-D quadric SLAM system is constructed which integrates the proposed constraint in object reconstruction, frontend pose estimation, and BA.

   \item Comprehensive experimental results prove our findings and demonstrate that our system outperforms state-of-the-art (SOTA) methods on public datasets.


\end{itemize}

\section{Related Works}
Object-based SLAM incorporates semantic objects as landmarks, enhancing the robustness and accuracy of the SLAM system while also providing precise semantic object maps for high-level tasks like object-based localization \cite{li2019semantic, deng2022object, wang2024goreloc} and object grasping \cite{wu2023object, wu2021object}.

DSP-SLAM \cite{wang2021dsp} adopts DeepSDF \cite{park2019deepsdf} as a shape embedding, combining sparse 3D points and deep shape priors to achieve dense reconstruction of semantic objects.
However, its reliance on prior shape information significantly limits its generality in reconstructing diverse objects.
In contrast, CubeSLAM \cite{yang2019cubeslam} utilizes cuboids inferred from 2D bounding boxes as object representations, 
achieving real-time 3D object detection and SLAM in both static and dynamic environments. 
QuadricSLAM \cite{nicholson2018quadricslam} uses dual quadrics to represent objects, reconstructing them from multiple views to obtain information about size, position, and orientation. 
Compared to cuboids, dual quadrics offer a closed-form correspondence between their projections in 2D images and the original 3D objects, providing better constraints for general objects.

In quadric-based SLAM, existing constraints are primarily derived from bbox \cite{nicholson2018quadricslam, ok2019robust, hosseinzadeh2019real, tian2021accurate, cao2022object, sunderhauf2017dual} and conics \cite{wang2023qiso, hosseinzadeh2019structure, wang2024voom, rubino20173d}.
The bbox-derived constraints calculate the coordinate distance or Intersection-Over-Union (IoU) between observed bounding boxes and those projected from dual quadrics.
In contrast, conic-derived constraints utilize more precise conics to describe objects in the 2D image space more accurately.
Recently, QISO-SLAM \cite{wang2023qiso} introduced precise segmentation contours to minimize the algebraic error \cite{chernov2004statistical, chojnacki2000fitting} between contours and reprojected conics, incorporating an outlier filtering strategy to enhance accuracy.

\section{Preliminary}
\label{sec:preliminary}
In standard form, a quadric surface $Q$ in projective space is defined by the equation:
\begin{equation}
   p^TQp=0
\label{eq_pointquadric}
\end{equation}
where $Q$ is a symmetric $4 \times 4$ matrix, and $p$ is a point represented in 4-dimensional homogeneous coordinates.

The projection of the quadric $Q$ onto the image plane is expressed as:
\begin{equation}
   \label{eq_quadricproject}
   C = H Q H^{T}
\end{equation}
where $C$ represents the $3\times3$ conic matrix, $H = K[R|t]$ is the $3\times4$ camera projection matrix composed of intrinsic matrix $K$ and extrinsic parameters $[R|t]$.

In its dual form, a dual quadric $Q^*$ is defined by its tangent planes $\pi$ as:
\begin{equation}
   \label{eq_dualquadric}
   \pi^TQ^*\pi = 0
\end{equation}
where $\pi$ is a 4-dimensional homogeneous vector representing a plane in 3D place.



$Q^*$ can represent both closed surfaces, such as ellipsoids, and non-closed surfaces, like hyperboloids \cite{hartley2003multiple}.
However, since closed surfaces are suitable for object landmark representation,
we use the constrained quadric to ensure that the surface is an ellipsoid.
Following the approach in \cite{rubino20173d}, the constrained dual quadric can be parameterized as follows:
\begin{equation}
   Q^*=Z\breve{Q^*}Z^T
\end{equation}
where $\breve{Q^*}$ represents an ellipsoid centered at the origin, 
$Z$ is a transformation matrix in homogeneous coordinates, accounting for an arbitrary rotation and translation. 
Specifically, $Z$ and $\breve{Q^*}$ are defined as:
\begin{equation}
Z = 
\begin{pmatrix} 
R(\theta) & t \\
0_3^\top & 1 
\end{pmatrix}
\quad \text{and} \quad
\check{Q^*} = 
\begin{pmatrix}
s_1^2 & 0 & 0 & 0 \\
0 & s_2^2 & 0 & 0 \\
0 & 0 & s_3^2 & 0 \\
0 & 0 & 0 & -1
\end{pmatrix}
\end{equation}
where \(t = (t_1, t_2, t_3)\) represents the translation vector of the ellipsoid centroid, 
\(R(\theta)\) is a rotation matrix defined by the angles \(\theta = (\theta_1, \theta_2, \theta_3)\), 
and \(s = (s_1, s_2, s_3)\) represents the lengths of three semi-axes of the ellipsoid.

\section{Convex Hull-based Algebraic Constraint}
\subsection{Existing Constraints for Quadric-based SLAM}

In Quadric SLAM, constraints for a quadric or dual quadric are dervied from various constraint primitives based on object detection and segmentation results. 
Common primitives include bbox, conic, and segmentation contour.
\subsubsection{Bbox-dervied constraint}
Bbox is the most commonly used primitive due to its simplicity and alignment with 2D object detectors \cite{nicholson2018quadricslam}. 
Constraints for bbox primitives are typically based on an overlap-based error term, which maximizes the overlapped area between the projected $C$ and the observed bounding box $b$:
\begin{equation}
   \mathcal{L}_o = -\zeta(C, b)
   \label{eq_overlap}
\end{equation}
where $\zeta$ is a function that quantifies the overlapping area. Commom implementations of $\zeta$ include IoU \cite{hosseinzadeh2019real} or pixel coordinate distance \cite{nicholson2018quadricslam}.

However, bbox only roughly approximates the object and the overlap-based error term is calculated as a scalar, which provides only a weak gradient for optimization.

\subsubsection{Conic-derived constraint}
Conic represents the projection of $Q$ onto the image plane (\Refeq{eq_quadricproject}), imposing a stronger constraint than bbox.
The distribution-based constraint \cite{cao2022object} constructs distribution distances by interpreting conics as 2D Gaussian distributions, enabling more accurate gradient computation:
\begin{equation}
   \mathcal{L}_d = \|\mu_z - \mu_c\|_2^2 + \text{Tr}(\Sigma_z + \Sigma_c - 2(\Sigma_z^{1 \over 2} \Sigma_c \Sigma_z^{1 \over 2})^{1 \over 2}),
\end{equation}
where \( \mu_z, \mu_c \) are the centers of the observed and predicted ellipses, and \( \Sigma_z, \Sigma_c \) are their respective covariance matrices encoding the shapes and orientations of the ellipses. 

A conic can be obtained either by fitting the largest inscribed ellipse within the bounding box \cite{rubino20173d} or by applying least squares fitting to contour points through instance segmentation \cite{wang2024voom}.
However, the former remains imprecise, while the latter is computationally nontrivial.

\begin{figure}[t]
   \raggedright
   \subfigure[The conic (red) derived from object contour (green)] 
   {
       \label{fig_conic_from_contours}
       \includegraphics[width=0.45\linewidth]{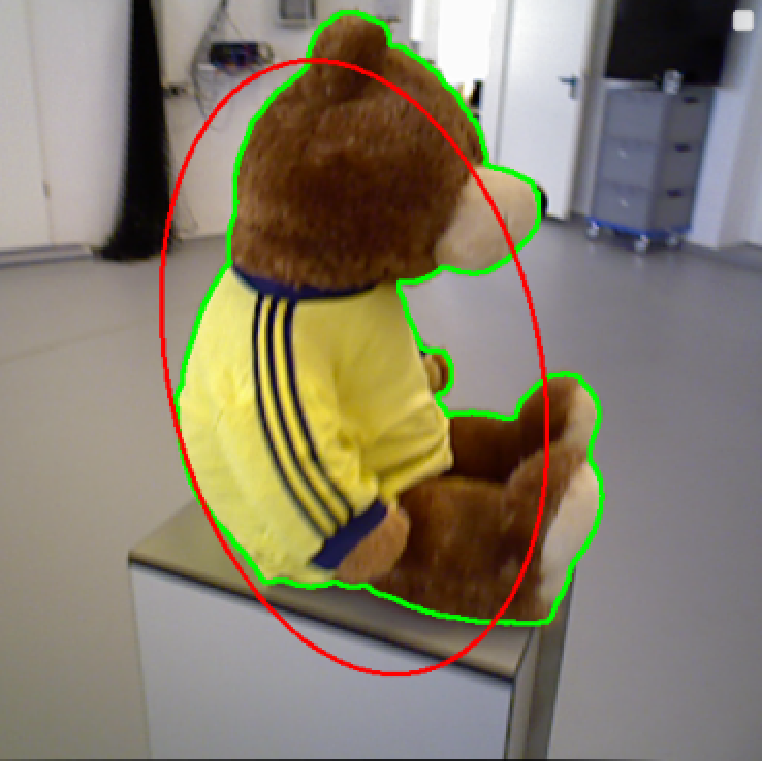}
   }
   \hfill
   \subfigure[The conic (red) derived from the convex hull (green).]
   {
       \label{fig_conic_from_convex_hull}
       \includegraphics[width=0.45\linewidth]{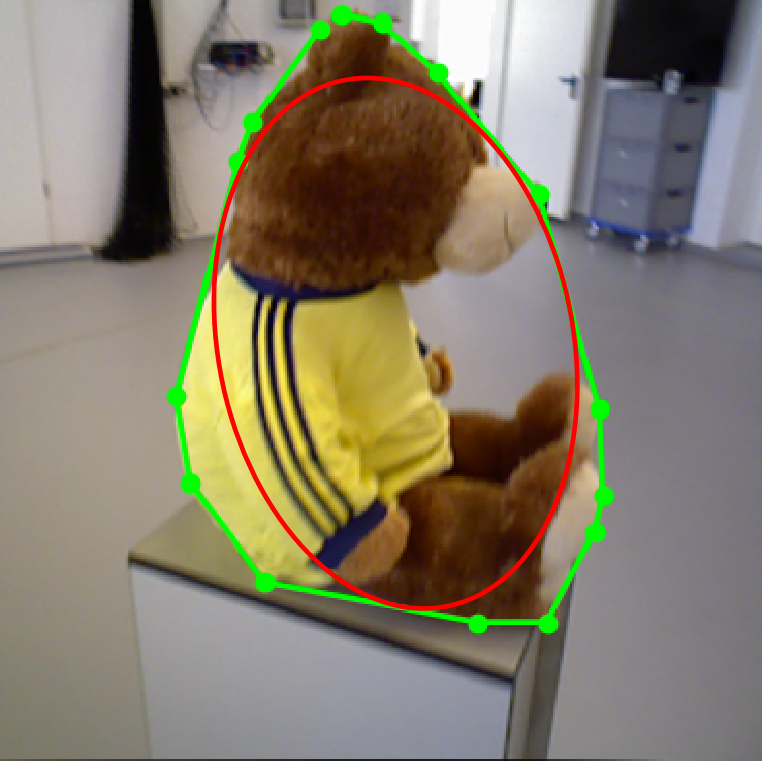}
   }
   \caption{
      Comparison of Conic Fitting from Contour and Convex Hull.
   }
   \label{fig_conic_fit}
   \vspace{-0.6cm}
\end{figure}

\subsubsection{Contour-derived constraint}
Contours provide an precise representation of the projected boundary of an object.
However, due to the approximation of objects by dual quadrics, the reprojected dual quadric (\ie conics) may misalign with segmented contours, particularly for objects with irregular or complex shapes.


To address this, QISO-SLAM \cite{wang2023qiso} adopts the point-based algebraic constraint (\Refeq{eq_pointquadric} and \Refeq{eq_quadricproject}) and introduces a contour outlier filtering strategy based on covariance-based maximum likelihood estimation.
However, it employs a fixed threshold derived from $\chi ^2$ testing to exclude contour points whose distances to the projected conic exceed the threshold.
This approach can hinder convergence, especially in the early stages of optimization.

\subsection{Convex Hull-based Algebraic Constraint}
\label{sec:ProposedConstraint}
To address the limitations of existing constraints, we propose a simple yet effective plane-based algebraic constraint (\Refeq{eq_dualquadric}) that incorporates convex hulls for precise dual quadric reconstruction and pose optimization:
\begin{equation}
   \mathcal{L}_{a^*} =\sum_{i} \left| \pi_{i}^TQ^*\pi_{i} \right|
   \label{eq_algebraic_dual_error}
\end{equation}
\begin{equation}
   \pi_{i} = H^Tl_{i}
   \label{eq_plane_from_line}
\end{equation}
where, $l_i$ represents a boundary segment of convex hulls as a 3-dimensional homogeneous vector, and \( H \) is the camera projection matrix.

This constraint leverages the precise contour information provided by instance segmentation. 
The convex hull of the contour is extracted to eliminate concave contour segments that could introduce inconsistencies in the dual quadric constraints across multiple views. 
To further stabilize optimization, we simplify the convex hull by removing excessive segments.

Compared to contour-based constraints, the convex hull retains the convex segments, providing a more precise constraint for the dual quadric, as demonstrated in \Reffig{fig_conic_fit}, where the reprojected dual conic optimized using the contour-based constraint deviates from the general shape of the object due to inconsistencies caused by concave contour segments across multiple views.

Additionally, based on multi-view geometry principles, a 2D line segment can uniquely project onto an infinite plane in 3D space. This equivalence between 2D and 3D formulations helps avoid the limitations of 2D points, which can represent any point along the corresponding ray, ensuring a more stable and precise constraint.

For implementation, we employ the Quick-Hull algorithm \cite{barber1996quickhull} to extract the convex hull from the segmented contours, and simplify it using the Douglas-Peucker line simplification method \cite{saalfeld1999topologically}.

\begin{figure*}[ht]
   \centering
   \includegraphics[width=17.5cm]{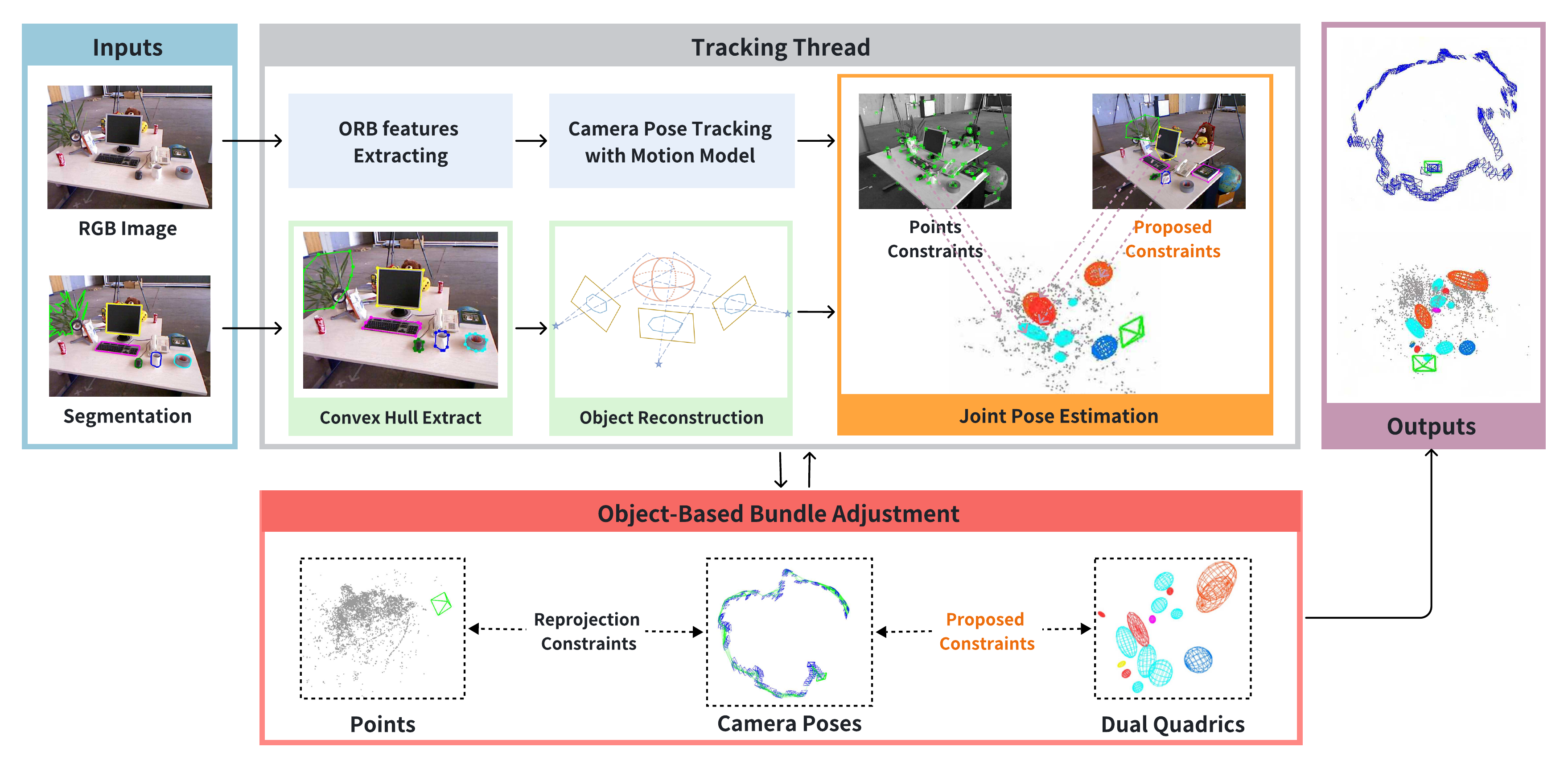}
   \vspace{-0.3cm}
   \caption{The proposed object SLAM system framework}
   \vspace{-0.4cm}
   \label{fig_slam}
\end{figure*}

\section{Integration into Quadric SLAM}
\label{sec:geometricslam}
In this section, we propose a quadric SLAM system based on ORB-SLAM2 \cite{mur2017orb}, integrating the proposed constraint into object reconstruction, frontend pose estimation, and backend BA.
The system is illustrated in \Reffig{fig_slam}, supporting both monocular and RGB-D mode.

We use YOLOv8 \cite{Jocher_Ultralytics_YOLO_2023} for 2D object detection and SAM \cite{kirillov2023segment} for offline accurate contour extraction.
In the tracking thread, the camera pose is initially estimated using the motion model, 
followed by refinement through Joint Pose Estimation (\Refsec{sec_joint_estimation}), which incorporates both point constraints and the proposed constraints.
Object reconstruction and data association are also performed in the tracking thread, as detailed in \Refsec{sec_reconstruct}.
In the backend, we perform Object-Based Bundle Adjustment (BA) to jointly optimize camera poses, dual quadrics, and feature points, as detailed in \Refsec{sec_ba}.
In the system, camera poses are represented as $X=\left\{x_{i} \in S E(3)\right\}$, feature points are $P=\left\{p_{j} \in \mathbb{R}^{3}\right\}$, and dual quadrics are $\mathbb{Q}=\left\{q_{k} \right\}$, where $q_k$ is represented by $Q^{*}$.  
The constraint includes the proposed convex hull-based algebraic constraint $\mathcal{L}_{a^*}$ and feature point reprojection constraint $\mathcal{L}_{p}$.

\subsection{Object Reconstruction and Data Association}
\label{sec_reconstruct}


Following \cite{song2022scale}, we employ the 3D oriented bounding box (OBB) for the rapid initialization of the dual quadric. 
After collecting multi-view observations \cite{nicholson2018quadricslam}, the observed convex hulls are utilized to refine the dual quadrics, formulated as:
\begin{equation}
   \label{eq_reconstruct}
   \left\{ q_k^* \in \mathbb{Q} \right\} = \underset{q_k}{\arg\min} \sum_{i} \mathcal{L}_{a^*}^i
\end{equation}
where \( i \) denotes the index of the observations.

We adopt the joint data association method proposed by \cite{wu2023object},
which integrates IoU-based, nonparametric, and t-test data association strategies.


\subsection{Joint Pose Estimation}
\label{sec_joint_estimation}


In the ORB-SLAM2 framework, tracking is highly sensitive to environmental changes and susceptible to failure in low-texture or dynamic environments.
To improve the robustness of the system, we propose a Joint Pose Estimation method 
that integrates both object information and point information for reliable pose estimation.
The optimization problem is formulated as:



\begin{equation}
   \label{eq_jpe}
   \begin{array}{l}
      X^{*}=\underset{\{X\}}{\arg \min }\left\{\sum_{i}\left\|\mathcal{L}_{p_i}\right\| + \sum_{j}\left\|\mathcal{L}_{a^*}\right\| \right\}
   \end{array}
\end{equation}
where $i$ and $j$ denote the indices of feature points and dual quadrics respectively.

\subsection{Bundle Adjustment}
\label{sec_ba}
In the backend BA, the joint optimization problem is formulated as:

\begin{equation}
   \label{eq_opt}
   \begin{array}{l}
      X^{*}, P^{*}, \mathbb{Q}^{*}=\underset{\{X, P, \mathbb{Q}\}}{\arg \min }\left\{\sum_{i, j}\left\|mathcal{L}_{p_j}^i\right\|_{\Omega_{i j}}^{2}\right. \\
      \left. +\sum_{i, k}\left\|\mathcal{L}_{a^*}^i\right\|_{\Omega_{i k}}^{2}\right\}
   \end{array}
\end{equation}
where $\Omega$ is the covariance matrix of different error measurements for Mahalanobis norm,
and $i$, $j$ and $k$ denote the indices of camera poses, feature points and dual quadrics respectively. 
This nonlinear least squares problem is efficiently solved using the Levenberg-Marquardt algorithm.

\section{Experiments}

We aim to address four research questions to evaluate the effectiveness of the proposed constraints in SLAM systems. 

Q1: Does our method outperform existing Quadric-based SLAM methods? 

Q2: Is our proposed constraint better than other constraints?

Q3: How is the convex hull better than contour as the constraint primitive?

Q4: Where to apply the proposed constraints in a SLAM system?

Experiments in \Refsec{sec_acc} correspond to Q1, focusing on comparing the performance of the proposed constraint with existing methods. 
\Refsec{sec_ablation} includes experiments addressing Q2, Q3, and Q4, conducting ablation studies on different constraints to assess their impact on mapping and localization accuracy and analyzing the integration of the proposed constraints into various components of the SLAM system. 


\begin{table*}[t]
   \centering
   \caption{Comparison of Pose Error on TUM RGB-D and ICL-NUIM Datasets}
   \vspace{-0.2cm}
   \begin{threeparttable}
      \setlength{\tabcolsep}{4pt}  
      \begin{tabular}{@{}c@{\hskip 0.10in}c@{\hskip 0.10in}c@{\hskip 0.10in}c@{\hskip 0.10in}c@{\hskip 0.10in}c@{\hskip 0.10in}c@{\hskip 0.10in}c@{\hskip 0.10in}c@{\hskip 0.10in}c@{\hskip 0.10in}c@{\hskip 0.13in}c@{}}       
         \toprule
         Method & fr1-desk & fr2-desk & fr2-person & fr2-dishes & fr3-office & fr3-teddy & office-traj0 & office-traj1 & office-traj2 & office-traj3 & Average\\ \midrule
         ORB-SLAM2 (mono)  & 0.0146 & 0.0096 & 0.0073  & - & 0.0141 & 0.0505 & 0.0521 & - & 0.0326 & - & 0.0258 \\
         OA-SLAM (mono)    & 0.0145 & 0.0092 & 0.0078  & - & 0.0128 & 0.0577 & 0.0517 & - & 0.0308 & - & 0.0264 \\
         OA-SLAM-BA (mono)    & 0.0150 & 0.0104 & 0.0073  & - & 0.0214 & 0.0759 & 0.0528 & - & 0.0375 & - & 0.0315 \\
         QuadricSLAM† (mono)   & 0.0167 & 0.0124 & -  & - & 0.0230 & - & - & - & - & - & - \\
         \cite{song2022scale} (mono)   & 0.0144 & 0.0087 & 0.0072  & - & 0.0177 & 0.0546 & 0.0560 & - & 0.0238 & - & 0.0261 \\
         Ours (mono)       & \textbf{0.0142} & \textbf{0.0071} & \textbf{0.0070}  & - & \textbf{0.0104} & \textbf{0.0223} & \textbf{0.0442} & - & \textbf{0.0197} & - & \textbf{0.0178} \\ \midrule
         
         ORB-SLAM2 (RGB-D) & 0.0171 & 0.0075 & 0.0065 & 0.0368 & 0.0107 & 0.0168 & 0.0260 & 0.0467 & 0.0097 & 0.0712 & 0.0249 \\
         QISO-SLAM† (RGB-D)& 0.0166 & 0.0099 & -  & - & 0.0112 & - & - & - & - & - & - \\
         VOOM (RGB-D)      & 0.0179 & 0.0074 & 0.0062 & 0.0202 & 0.0103 & 0.0160 & 0.0240 & \textbf{0.0373} & 0.0108 & 0.0689 & 0.0219 \\
         Ours (RGB-D)      & \textbf{0.0161} & \textbf{0.0068} & \textbf{0.0060} & \textbf{0.0125} & \textbf{0.0095} & \textbf{0.0145} & \textbf{0.0235} & 0.0385 & \textbf{0.0092} & \textbf{0.0638} & \textbf{0.0200} \\
         \bottomrule
      \end{tabular}
      \begin{tablenotes}
         \footnotesize
         \item † Results are directly reported from the respective papers. Best results are highlighted in \textbf{bold}. Due to the absence of stable visual features, all monocular methods listed in the table failed to estimate trajectories for the fr2-dishes, office-traj1, and office-traj3 sequences.
      \end{tablenotes}
   \end{threeparttable}
   \label{tb_traj}
\end{table*}

\subsection{Experimental Settings}
\subsubsection{Datasets and Baselines}


We evaluate the proposed constraint using the TUM RGB-D dataset \cite{sturm2012benchmark} and the synthesized ICL-NUIM dataset \cite{handa2014benchmark}.
Performance in both the monocular and RGB-D modes is evaluated. 
In the monocular mode, comparisons are made against OA-SLAM \cite{zins2022oa}, QuadricSLAM \cite{nicholson2018quadricslam}, Song \cite{song2022scale} and ORB-SLAM2 \cite{mur2017orb}.
Additionally, we include a variant of OA-SLAM with backend-optimization using objects (OA-SLAM-BA). 
In the RGB-D mode, we compare our method with ORB-SLAM2, QISO-SLAM \cite{wang2023qiso} and VOOM \cite{wang2024voom}.
All experiments are conducted with loop closure enabled and all methods share the same object detector.

\subsubsection{Metrics}
\label{sec_benchmark}

We adopt root-mean-square error (RMSE) to quantify the Absolute Trajectory Error (ATE) under scale alignment.

To evaluate mapping accuracy, we propose two metrics.
SIoU is an IoU-based metric that measures the overlap between the segmentation contour and the reprojected conic: 
\begin{equation} 
   \text{SIoU} = \frac{\text{Area}(I)}{\text{Area}(C) + \text{Area}(S) - \text{Area}(I)} 
\end{equation}
where $C$ denotes the reprojected conic, $S$ denotes the segmentation contour, and $I$ denotes the intersection between $C$ and $S$, which is computed by a geometry-based polygon clipping method \cite{de2000computational}.
The area of each polygon is calculated with the shoelace theorem \cite{lee2017shoelace}.

In addition to SIoU, we also use the mean tangent distance (MTD) to evaluate the mapping accuracy under the elimination of inaccuracies in non-convex contours.
The MTD is defined as:
\begin{equation}
   \label{eq_tangent}
   \text{MTD} = \frac{\sum_{i=1}^M \mathcal{L}_{a^*}^i}{\sum_{i=1}^M N_i},
\end{equation}
where $M$ is the number of observations, and $N_i$ denotes the number of line segments for the $i$-th observation.



\begin{figure*}[ht]
   \centering
   \begin{tabular}{ccc}
      \includegraphics[width=0.28\linewidth]{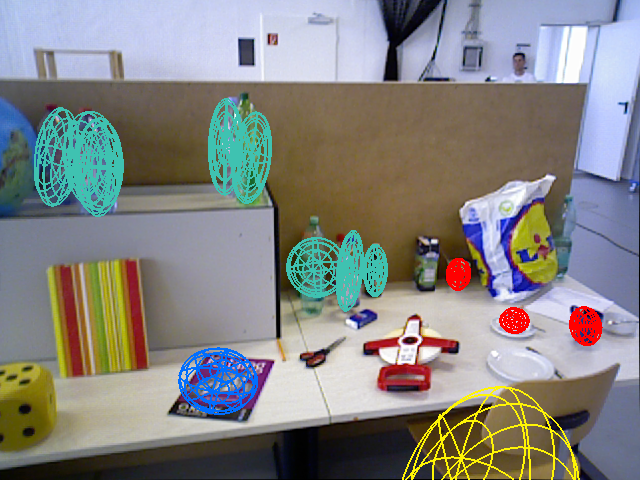} &
      \includegraphics[width=0.28\linewidth]{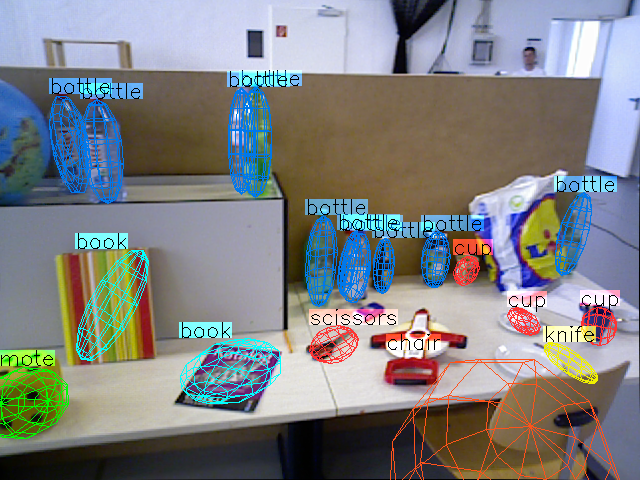} &
      \includegraphics[width=0.28\linewidth]{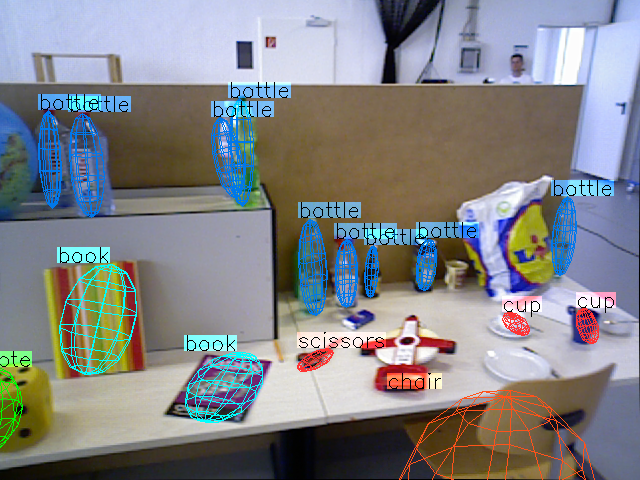} \\
      \includegraphics[width=0.28\linewidth]{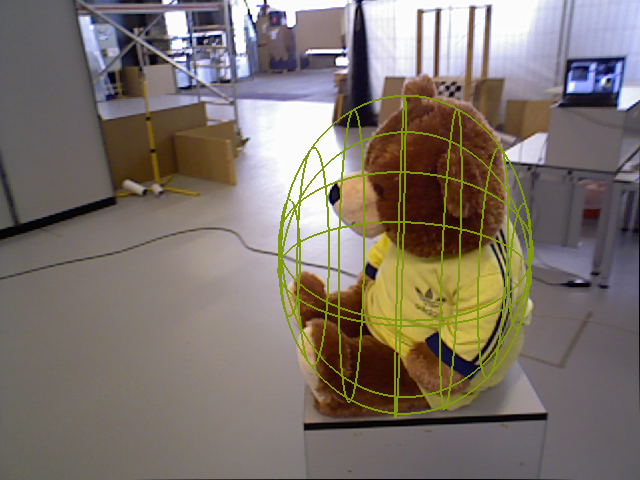} &
      \includegraphics[width=0.28\linewidth]{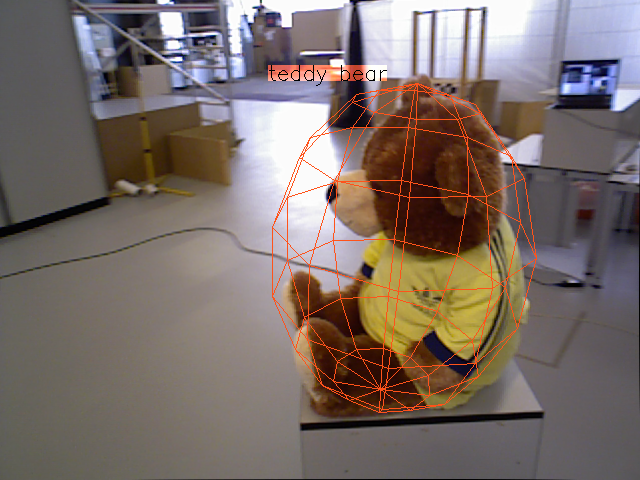} &
      \includegraphics[width=0.28\linewidth]{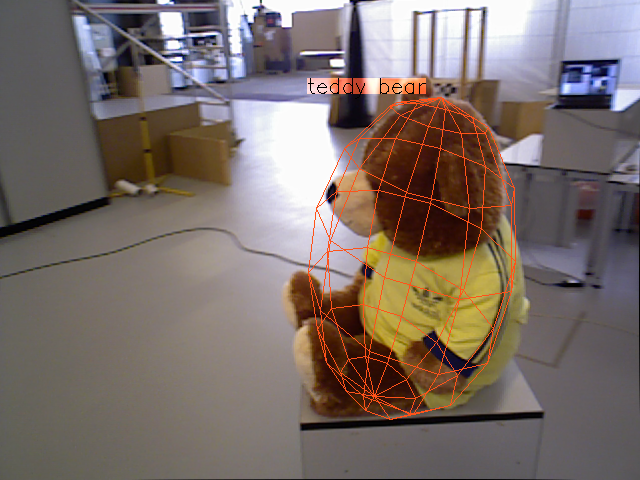} \\
      (a) OA-SLAM & (b) \cite{song2022scale} & (c) Ours
   \end{tabular}
   \caption{
      Qualitative comparison of mapping results on the TUM RGB-D dataset.
      The first column represents the fr3-office sequence, and the second column represents the fr3-teddy sequence. 
   }
   \label{fig_qualitative_comparison}
   \vspace{-0.5cm}
\end{figure*}

\subsection{Localization and Mapping}
\label{sec_acc}





\Reftab{tb_traj} compares the localization accuracy across multiple sequences from the TUM RGB-D and ICL-NUIM datasets, covering both monocular and RGB-D modes. 

In the monocular mode, our method achieves the highest localization accuracy across all tested sequences. 
This demonstrates the effectiveness of the proposed constraint in improving localization performance. 
Notably, in the fr3-teddy sequence, our method significantly outperforms other baselines, showcasing its adaptability in handling complex-shaped objects (as illustrated in \Reffig{fig_conic_fit}). 

In the RGB-D mode, our method outperforms baseline approaches in most sequences. 
This result highlights the generalizability of the proposed constraints. 

\begin{table}[tbp]
   \raggedright  
   \caption{Mapping Quality Evaluation}
   \vspace{-0.2cm}
   \label{tb_mapping}
   \begin{threeparttable}
     \setlength{\tabcolsep}{4pt}  
     \begin{tabular}{ccccccc}
       \toprule
       \multirow{2}{*}{\makecell[c]{sequence}} & \multicolumn{3}{c}{MTD ($10^{-3}$) $\downarrow$} & \multicolumn{3}{c}{SIoU $\uparrow$} \\
       \cmidrule(lr){2-4}  \cmidrule(lr){5-7} &OA-SLAM & \cite{song2022scale} & Ours &OA-SLAM & \cite{song2022scale} & Ours \\
       \midrule
       fr1-desk   & 7.817 & 11.679 & \textbf{6.385} & 0.678          & 0.653 & \textbf{0.719} \\
       fr2-desk   & 1.629 & 1.216 & \textbf{0.963} & 0.636           & 0.676 & \textbf{0.705} \\
       fr2-person & 2.454 & 2.716 & \textbf{0.712} & 0.642           & 0.655 & \textbf{0.700} \\
       fr3-office & 1.780 & 1.594 & \textbf{0.826} & 0.636           & 0.696 & \textbf{0.705} \\
       fr3-teddy  & 21.645 & 19.737 & \textbf{3.107} & 0.717         & 0.719 & \textbf{0.764} \\
       Average    & 5.907 & 6.171 & \textbf{2.005} & 0.662           & 0.680 & \textbf{0.719} \\
       \bottomrule
     \end{tabular}
     \begin{tablenotes}
         \footnotesize 
         \item Higher $\uparrow$ and lower $\downarrow$ indicate preferred metrics; 
         the \textbf{best} results are highlighted.
     \end{tablenotes}
   \end{threeparttable}
   \vspace{-0.6cm}
\end{table}

\begin{table*}[t]
   \centering
   \caption{Ablation Study for Common Constraints}
   \vspace{-0.2cm}
   \label{tb_ablation_constraint}
   \begin{threeparttable}
     \setlength{\tabcolsep}{4.5pt}  
     \begin{tabular}{lcccccccccccccccc}
       \toprule
       \multirow{3}{*}{\makecell[c]{Sequence}} & \multicolumn{8}{c}{Overlap} & \multicolumn{2}{c}{Distribution} & \multicolumn{2}{c}{Algebraic(point)} & \multicolumn{4}{c}{Algebraic(plane)} \\
       \cmidrule(lr){2-9} \cmidrule(lr){10-11} \cmidrule(lr){12-13} \cmidrule(lr){14-17}
       & \multicolumn{2}{c}{bbox\cite{nicholson2018quadricslam}} & \multicolumn{2}{c}{conic} & \multicolumn{2}{c}{contour} & \multicolumn{2}{c}{convex hull} & \multicolumn{2}{c}{conic\cite{zins2022oa}} & \multicolumn{2}{c}{contour\cite{wang2023qiso}} & \multicolumn{2}{c}{bbox\cite{sunderhauf2017dual}} & \multicolumn{2}{c}{convex hull} \\
       \cmidrule(lr){2-3} \cmidrule(lr){4-5} \cmidrule(lr){6-7} \cmidrule(lr){8-9} \cmidrule(lr){10-11} \cmidrule(lr){12-13} \cmidrule(lr){14-15} \cmidrule(lr){16-17}
       & ATE & SIoU & ATE & SIoU & ATE & SIoU & ATE & SIoU & ATE & SIoU & ATE & SIoU & ATE & SIoU & ATE & SIoU \\
       \midrule
       fr1-desk   & 1.456 & 0.647 & 1.442 & 0.621 & 1.512 & 0.691          & 1.472 & 0.695 & 1.501 & 0.614 & 1.458 & 0.606 & 1.473 & 0.610 & \textbf{1.414} & \textbf{0.719} \\
       fr2-desk   & 0.766 & 0.676 & 0.786 & 0.637 & 0.852 & \textbf{0.707} & 0.731 & 0.696 & 0.769 & 0.655 & 0.753 & 0.652 & 0.736 & 0.676 & \textbf{0.678} & 0.705 \\
       fr2-person & 0.778 & 0.653 & 0.717 & 0.635 & 0.814 & \textbf{0.701} & 0.741 & 0.699 & 0.773 & 0.653 & 0.843 & 0.624 & 0.775 & 0.643 & \textbf{0.688} & 0.700 \\
       fr3-office & 1.116 & 0.682 & 1.102 & 0.631 & 1.347 & 0.688          & 1.085 & 0.672 & 1.099 & 0.629 & 1.181 & 0.643 & 1.100 & 0.700 & \textbf{1.028} & \textbf{0.700} \\
       fr3-teddy  & 4.578 & 0.639 & 3.936 & 0.697 & -     & -              & 3.236 & 0.749 & 3.779 & 0.707 & 3.781 & 0.708 & 4.195 & 0.696 & \textbf{2.064} & \textbf{0.764} \\
       \midrule
       Average    & 1.739 & 0.660 & 1.597 & 0.644 & -     & -              & 1.453 & 0.702 & 1.586 & 0.652 & 1.615 & 0.646 & 1.656 & 0.665 & \textbf{1.174} & \textbf{0.717} \\
       \bottomrule
     \end{tabular}
     \begin{tablenotes}
         \footnotesize 
         \item The first row represents the error terms, and the second row represents the constraint primitives used in the study. Higher \textbf{SIoU} and lower \textbf{ATE} values (in centimeters) are preferred. Best results are highlighted in \textbf{bold}.
     \end{tablenotes}
   \end{threeparttable}
   \vspace{-0.3cm}
\end{table*}

\begin{table*}[t]
      \centering
      \caption{Ablation Study For Plane-based Algebraic Constraint }
      \vspace{-0.2cm}
      \label{tb_ablation_geometric_constraint}
      \begin{threeparttable}
        \setlength{\tabcolsep}{9pt}  
        \begin{tabular}{lcccccccccccccccc}
          \toprule
          \multirow{2}{*}{\makecell[c]{Sequence}} & \multicolumn{2}{c}{Contour(0)} & \multicolumn{2}{c}{Contour(3)} & \multicolumn{2}{c}{Contour(6)} & \multicolumn{2}{c}{ConvexHull(0)} & \multicolumn{2}{c}{ConvexHull(3)} & \multicolumn{2}{c}{ConvexHull(6)} \\
          \cmidrule(lr){2-3} \cmidrule(lr){4-5} \cmidrule(lr){6-7} \cmidrule(lr){8-9} \cmidrule(lr){10-11} \cmidrule(lr){12-13} 
          & ATE & SIoU & ATE & SIoU & ATE & SIoU & ATE & SIoU & ATE & SIoU & ATE & SIoU \\
          \midrule
          fr1-desk   & 1.482 & 0.519 & 1.444 & 0.643 & 1.505 & 0.584 & 1.436    & 0.715          & 1.414          & \textbf{0.719} & \textbf{1.403} & 0.674 \\
          fr2-desk   & 0.776 & 0.589 & 0.821 & 0.633 & 0.743 & 0.596 & 0.692    & \textbf{0.725} & \textbf{0.678} & 0.705          & 0.743 & 0.639 \\
          fr2-person & 0.782 & 0.601 & 0.736 & 0.624 & 0.768 & 0.626 & 0.718    & \textbf{0.722} & \textbf{0.688} & 0.700          & 0.730 & 0.658 \\
          fr3-office & 1.114 & 0.579 & 1.074 & 0.641 & 1.120 & 0.615 & 1.049    & \textbf{0.705} & \textbf{1.028} & 0.700          & 1.066 & 0.666 \\
          fr3-teddy  & 6.130 & 0.453 & 5.200 & 0.655 & 3.665 & 0.543 & 2.568    & \textbf{0.765} & \textbf{2.064} & 0.764          & 2.771 & 0.755 \\
          \midrule
          Average   & 2.062 & 0.548 & 1.855 & 0.639 & 1.560 & 0.593 & 1.293    & \textbf{0.726} & \textbf{1.174} & 0.717          & 1.343 & 0.678 \\
          \bottomrule
        \end{tabular}
        \begin{tablenotes}
            \footnotesize 
            \item Higher \textbf{SIoU} and lower \textbf{ATE} values (in centimeters) are preferred. Best results are highlighted in \textbf{bold}.
        \end{tablenotes}
      \end{threeparttable}
   \end{table*}


\Reftab{tb_mapping} presents the mapping quality evaluation in TUM RGB-D dataset. 
Our method outperforms OA-SLAM and \cite{song2022scale} in both SIoU and MTD metrics for most sequences, demonstrating that the proposed constraint effectively improves mapping quality by ensuring consistent alignment across multiple views.
Notably, in the fr3-teddy sequence, 
our method shows significant improvements over OA-SLAM and \cite{song2022scale}, showcasing its superior capability in handling complex-shaped objects.







\begin{figure*}
   \centering
   \subfigure[]
   {
       \label{fig_qualitative_proj}
       \includegraphics[width=0.23\linewidth]{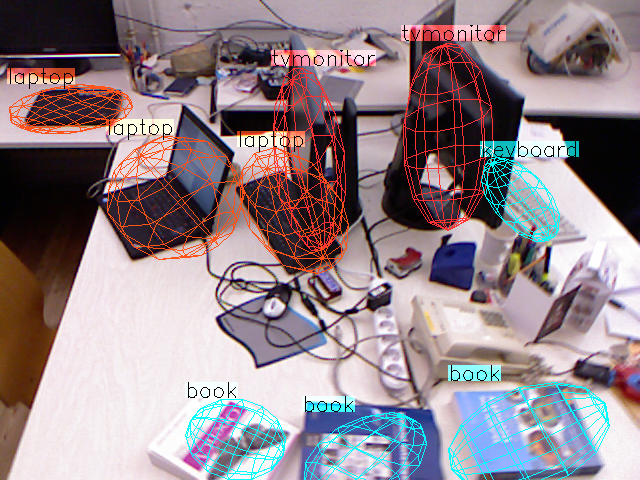}
       \includegraphics[width=0.23\linewidth]{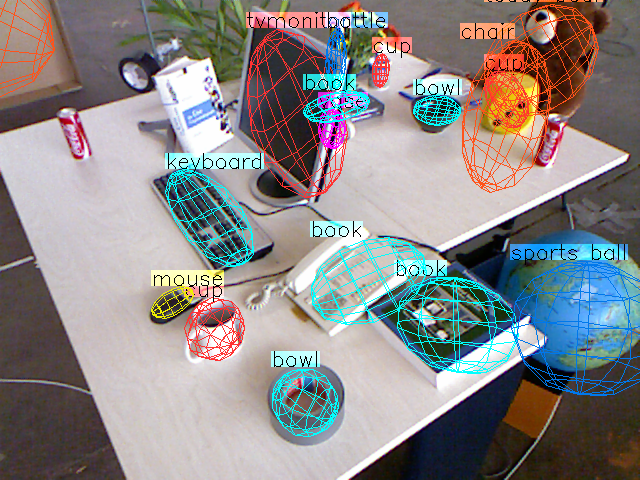}
       \includegraphics[width=0.23\linewidth]{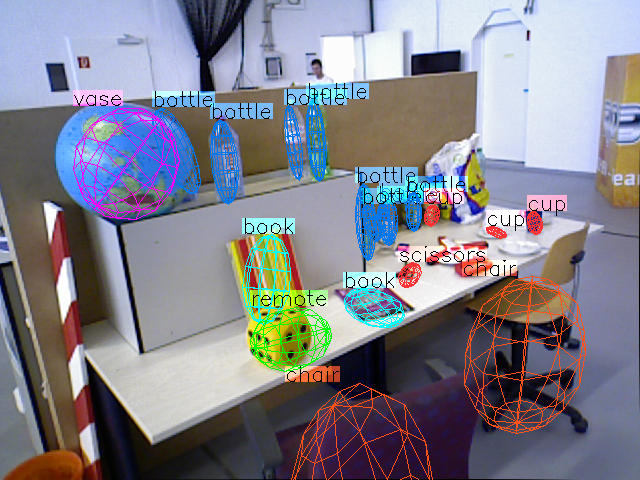}
       \includegraphics[width=0.23\linewidth]{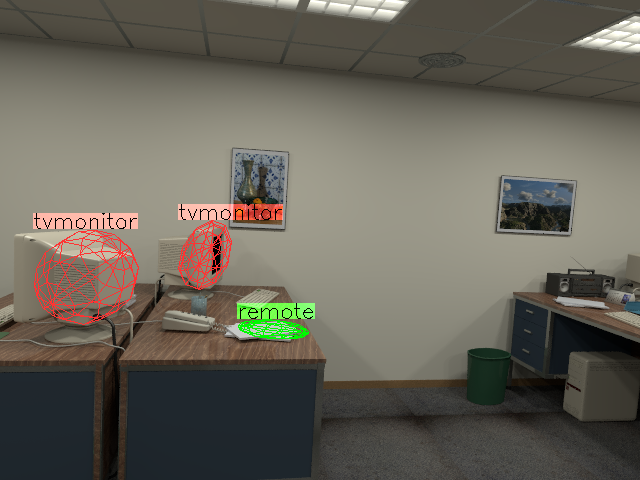}
   }
   \\
   \subfigure[]
   {
       \label{fig_qualitative_quadric}
       \includegraphics[width=0.23\linewidth]{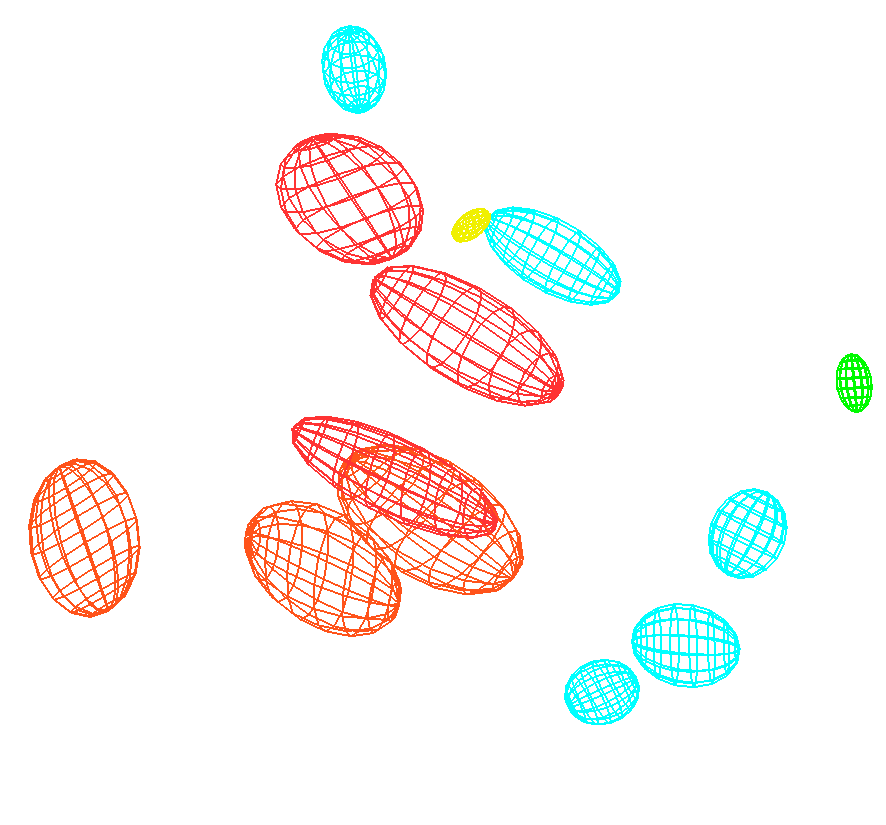}
       \includegraphics[width=0.23\linewidth]{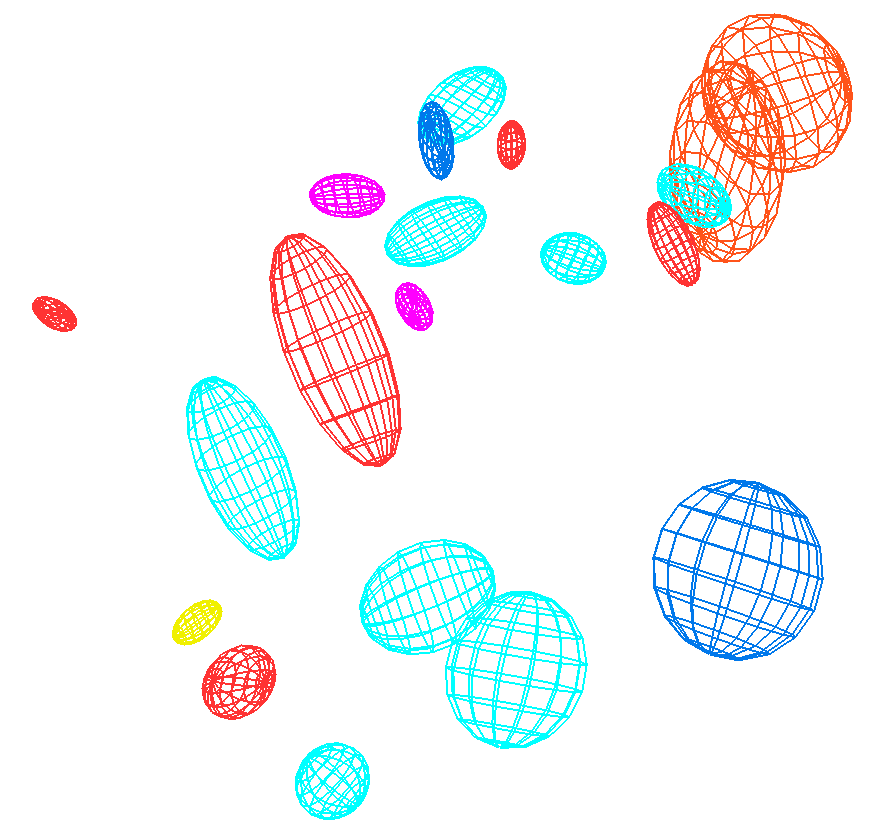}
       \includegraphics[width=0.23\linewidth]{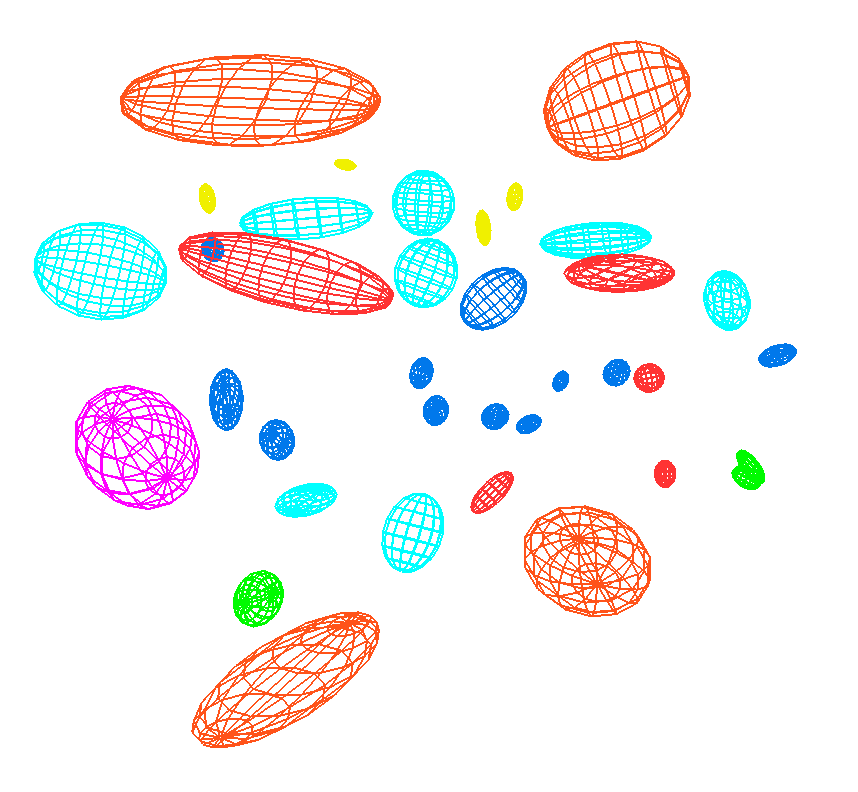}
       \includegraphics[width=0.23\linewidth]{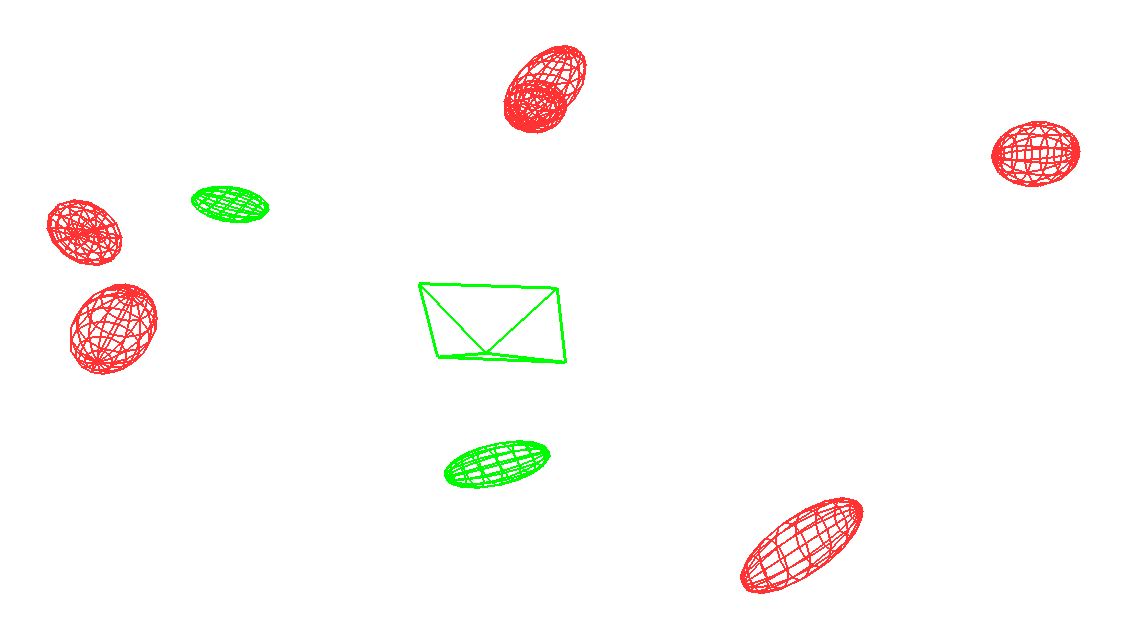}
   }
   \\
   \caption{The qualitative results of semantic mapping. 
   From left to right, columns are results on sequence TUM fr1\_desk, fr2\_desk, fr3\_desk and ICL NUIM office-traj2 respectively.
   (a) The 2-D projection of reconstructed quadrics on the image.
   (b) The object map built by our object SLAM.
   }
   \label{fig_qualitative}
   \vspace{-0.5cm}
\end{figure*}

\Reffig{fig_qualitative_comparison} provides a qualitative comparison of the mapping results of our method with OA-SLAM and \cite{song2022scale}.
The resulting dual quadrics are projected onto the same view.
Although all of these methods successfully reconstruct objects in the scene, the dual quadrics reconstructed by our method capture objects more accurately.
Additional qualitative results of the semantic mapping of our method are shown in \Reffig{fig_qualitative}.
\Reffig{fig_qualitative_proj} presents the projections of dual quadrics.
\Reffig{fig_qualitative_quadric} displays the object maps generated by our method.
It should be noted that the number of mapped objects can be limited by the performance of the adopted object detector.

Overall, the experimental results demonstrate that our method surpasses existing baseline methods in both localization accuracy and mapping quality, confirming the positive answer of Q1.

\subsection{Ablation Study}
\label{sec_ablation}

\subsubsection{Evaluation of Convex Hull-based Algebraic Constraint Against Other Constraints (Q2)}

We conducted ablation experiments with different combinations of constraint primitives and error terms in the system. 
\Reftab{tb_ablation_constraint} shows the performance of overlap-based, distribution-based, and algebraic constraints with various primitives on the TUM RGB-D dataset, using ATE and SIoU as evaluation metrics. 

Overall, the convex hull-based algebraic constraint demonstrates superior object alignment and consistency across diverse scenarios compared to other constraints. 


As for constraint primitives, convex hull outperforms all other primitives in the overlap-based error term. In the plane-based algebraic error term, convex hull also demonstrates better performance than bbox.












\subsubsection{Evaluating How Convex Hull Is Better than Contour as the Constraint Primitive (Q3)}




We compare plane-based algebraic constraints based on contour and convex hull across various levels of detail. 
The results are presented in \Reftab{tb_ablation_geometric_constraint}. 
The numbers indicate the degree of simplification applied to the primitive, with higher values corresponding to more aggressive simplification.

The results show that convex hull demonstrated superior performance than contour in general. 
ConvexHull(3) achieved the best average ATE for localization, while ConvexHull(0) attained the highest average SIoU for mapping. 
Although contour-based constraints show some improvement after simplification, their performance still fell short of those of convex hull.


\begin{table}[tbp]
   \raggedright  
   \setlength\tabcolsep{2.5pt}
   \begin{center}
       \begin{threeparttable}
       \caption{Ablation Study for System Integration}
       \label{tb_ablation}
       \begin{tabular}{@{}cccccccccccccccccccc@{}}
       \toprule
       JPE\tnote{1} & obj\_BA\tnote{2}        & fr1-desk             & fr2-desk        & fr2-person      & fr3-office         & fr3-teddy  &Average \\ \midrule
                            &                 & 0.0145               & 0.0082          & 0.0072          & 0.0167    & 0.0390     & 0.0171\\
       $\surd$              &                 & 0.0144               & 0.0074          & 0.0076          & 0.0104    & 0.0317     & 0.0143\\
                            & $\surd$         & \textbf{0.0142}      & 0.0069          & \textbf{0.0070} & 0.0103    & 0.0240     & 0.0125\\
       $\surd$              & $\surd$         & 0.0143               & \textbf{0.0068} & 0.0074          &\textbf{0.0098}    &\textbf{0.0227}     &\textbf{0.0122}\\ \bottomrule

   \end{tabular}
   \begin{tablenotes}
       \footnotesize 
       \item[1] Joint Pose Estimation.
       $^2$ Object-Based Bundle Adjustment.
   \end{tablenotes}
   \end{threeparttable}
   \end{center}
   \vspace{-0.7cm}
\end{table}

\subsubsection{Evaluation of Convex Hull-based Algebraic Constraints in SLAM System (Q4)}

As described in \Refsec{sec_joint_estimation} and \Refsec{sec_ba}, we apply the proposed constraint in both the front-end pose estimation and the back-end BA. 
To further evaluate its effectiveness within a SLAM system, we conduct ablation studies to analyze the impact of the proposed constraint in each system component.
We assess the localization accuracy across four configurations: a baseline without the proposed constraints, w/ Joint Pose Estimation (JPE), w/ Object BA (obj\_BA), and the Integration of both. 

The results, shown in \Reftab{tb_ablation}, indicate that the integration of JPE and obj\_BA achieves the best average performance. 
However, in the fr2-person sequence which contains moving figures, both JPE and the integration introduce errors.
This is because JPE relies on single-frame observations for optimization, making it sensitive to dynamic environments.
In contrast, obj\_BA leverages global information for optimization, effectively mitigating the impact of dynamic environments and improving localization accuracy.

\section{CONCLUSIONS}


In this paper, we propose a simple yet effective convex hull-based algebraic constraint for quadric SLAM. 
This approach leverages precise contour information from instance segmentation. 
It also addresses inconsistencies caused by complex object shapes in multi-view observations, thereby enhancing the alignment between object observations and dual quadrics.

By integrating this constraint into object reconstruction, pose estimation, and bundle adjustment within a SLAM system, we achieve significant improvements in both localization and mapping performance.

For future work, we plan to extend our approach to address relocalization challenges in monocular visual SLAM. 
Additionally, we aim to explore the integration of object information with structural elements to further enhance the robustness and adaptability of SLAM in complex and large-scale environments.

\bibliographystyle{IEEEtran}
\bibliography{reference}

\begin{thebibliography}{10}
\providecommand{\url}[1]{#1}
\csname url@rmstyle\endcsname
\providecommand{\newblock}{\relax}
\providecommand{\bibinfo}[2]{#2}
\providecommand\BIBentrySTDinterwordspacing{\spaceskip=0pt\relax}
\providecommand\BIBentryALTinterwordstretchfactor{4}
\providecommand\BIBentryALTinterwordspacing{\spaceskip=\fontdimen2\font plus
\BIBentryALTinterwordstretchfactor\fontdimen3\font minus \fontdimen4\font\relax}
\providecommand\BIBforeignlanguage[2]{{%
\expandafter\ifx\csname l@#1\endcsname\relax
\typeout{** WARNING: IEEEtran.bst: No hyphenation pattern has been}%
\typeout{** loaded for the language `#1'. Using the pattern for}%
\typeout{** the default language instead.}%
\else
\language=\csname l@#1\endcsname
\fi
#2}}

\bibitem{nicholson2018quadricslam}
L.~Nicholson, M.~Milford, and N.~S{\"u}nderhauf, ``Quadricslam: Dual quadrics from object detections as landmarks in object-oriented slam,'' \emph{IEEE Robotics and Automation Letters}, vol.~4, no.~1, pp. 1--8, 2018.

\bibitem{yang2019cubeslam}
S.~Yang and S.~Scherer, ``Cubeslam: Monocular 3-d object slam,'' \emph{IEEE Transactions on Robotics}, vol.~35, no.~4, pp. 925--938, 2019.

\bibitem{wu2020eao}
Y.~Wu, Y.~Zhang, D.~Zhu, Y.~Feng, S.~Coleman, and D.~Kerr, ``Eao-slam: Monocular semi-dense object slam based on ensemble data association,'' in \emph{2020 IEEE/RSJ International Conference on Intelligent Robots and Systems (IROS)}.\hskip 1em plus 0.5em minus 0.4em\relax IEEE, 2020, pp. 4966--4973.

\bibitem{zins2022oa}
M.~Zins, G.~Simon, and M.-O. Berger, ``Oa-slam: Leveraging objects for camera relocalization in visual slam,'' in \emph{2022 IEEE international symposium on mixed and augmented reality (ISMAR)}.\hskip 1em plus 0.5em minus 0.4em\relax IEEE, 2022, pp. 720--728.

\bibitem{wang2023qiso}
Y.~Wang, B.~Xu, W.~Fan, and C.~Xiang, ``Qiso-slam: Object-oriented slam using dual quadrics as landmarks based on instance segmentation,'' \emph{IEEE Robotics and Automation Letters}, vol.~8, no.~4, pp. 2253--2260, 2023.

\bibitem{hartley2003multiple}
R.~Hartley and A.~Zisserman, \emph{Multiple view geometry in computer vision}.\hskip 1em plus 0.5em minus 0.4em\relax Cambridge university press, 2003.

\bibitem{ok2019robust}
K.~Ok, K.~Liu, K.~Frey, J.~P. How, and N.~Roy, ``Robust object-based slam for high-speed autonomous navigation,'' in \emph{2019 International Conference on Robotics and Automation (ICRA)}.\hskip 1em plus 0.5em minus 0.4em\relax IEEE, 2019, pp. 669--675.

\bibitem{hosseinzadeh2019real}
M.~Hosseinzadeh, K.~Li, Y.~Latif, and I.~Reid, ``Real-time monocular object-model aware sparse slam,'' in \emph{2019 international conference on robotics and automation (ICRA)}.\hskip 1em plus 0.5em minus 0.4em\relax IEEE, 2019, pp. 7123--7129.

\bibitem{tian2021accurate}
R.~Tian, Y.~Zhang, Y.~Feng, L.~Yang, Z.~Cao, S.~Coleman, and D.~Kerr, ``Accurate and robust object slam with 3d quadric landmark reconstruction in outdoors,'' \emph{IEEE Robotics and Automation Letters}, vol.~7, no.~2, pp. 1534--1541, 2021.

\bibitem{sunderhauf2017dual}
N.~S{\"u}nderhauf and M.~Milford, ``Dual quadrics from object detection boundingboxes as landmark representations in slam,'' \emph{arXiv preprint arXiv:1708.00965}, 2017.

\bibitem{li2019semantic}
J.~Li, D.~Meger, and G.~Dudek, ``Semantic mapping for view-invariant relocalization,'' in \emph{2019 International Conference on Robotics and Automation (ICRA)}.\hskip 1em plus 0.5em minus 0.4em\relax IEEE, 2019, pp. 7108--7115.

\bibitem{deng2022object}
Z.~Deng, Y.~Zhang, Y.~Wu, Z.~Ge, X.~Hu, and W.~Sun, ``Object-plane co-represented and graph propagation-based semantic descriptor for relocalization,'' \emph{IEEE Robotics and Automation Letters}, vol.~7, no.~4, pp. 11\,023--11\,030, 2022.

\bibitem{wang2024goreloc}
Y.~Wang, C.~Jiang, and X.~Chen, ``Goreloc: Graph-based object-level relocalization for visual slam,'' \emph{IEEE Robotics and Automation Letters}, 2024.

\bibitem{wu2023object}
Y.~Wu, Y.~Zhang, D.~Zhu, Z.~Deng, W.~Sun, X.~Chen, and J.~Zhang, ``An object slam framework for association, mapping, and high-level tasks,'' \emph{IEEE Transactions on Robotics}, vol.~39, no.~4, pp. 2912--2932, 2023.

\bibitem{wu2021object}
Y.~Wu, Y.~Zhang, D.~Zhu, X.~Chen, S.~Coleman, W.~Sun, X.~Hu, and Z.~Deng, ``Object slam-based active mapping and robotic grasping,'' in \emph{2021 International Conference on 3D Vision (3DV)}.\hskip 1em plus 0.5em minus 0.4em\relax IEEE, 2021, pp. 1372--1381.

\bibitem{wang2021dsp}
J.~Wang, M.~R{\"u}nz, and L.~Agapito, ``Dsp-slam: Object oriented slam with deep shape priors,'' in \emph{2021 International Conference on 3D Vision (3DV)}.\hskip 1em plus 0.5em minus 0.4em\relax IEEE, 2021, pp. 1362--1371.

\bibitem{park2019deepsdf}
J.~J. Park, P.~Florence, J.~Straub, R.~Newcombe, and S.~Lovegrove, ``Deepsdf: Learning continuous signed distance functions for shape representation,'' in \emph{Proceedings of the IEEE/CVF conference on computer vision and pattern recognition}, 2019, pp. 165--174.

\bibitem{cao2022object}
Z.~Cao, Y.~Zhang, R.~Tian, R.~Ma, X.~Hu, S.~Coleman, and D.~Kerr, ``Object-aware slam based on efficient quadric initialization and joint data association,'' \emph{IEEE Robotics and Automation Letters}, vol.~7, no.~4, pp. 9802--9809, 2022.

\bibitem{hosseinzadeh2019structure}
M.~Hosseinzadeh, Y.~Latif, T.~Pham, N.~Suenderhauf, and I.~Reid, ``Structure aware slam using quadrics and planes,'' in \emph{Computer Vision--ACCV 2018: 14th Asian Conference on Computer Vision, Perth, Australia, December 2--6, 2018, Revised Selected Papers, Part III 14}.\hskip 1em plus 0.5em minus 0.4em\relax Springer, 2019, pp. 410--426.

\bibitem{wang2024voom}
Y.~Wang, C.~Jiang, and X.~Chen, ``Voom: Robust visual object odometry and mapping using hierarchical landmarks,'' \emph{arXiv preprint arXiv:2402.13609}, 2024.

\bibitem{rubino20173d}
C.~Rubino, M.~Crocco, and A.~Del~Bue, ``3d object localisation from multi-view image detections,'' \emph{IEEE transactions on pattern analysis and machine intelligence}, vol.~40, no.~6, pp. 1281--1294, 2017.

\bibitem{chernov2004statistical}
N.~Chernov and C.~Lesort, ``Statistical efficiency of curve fitting algorithms,'' \emph{Computational statistics \& data analysis}, vol.~47, no.~4, pp. 713--728, 2004.

\bibitem{chojnacki2000fitting}
W.~Chojnacki, M.~J. Brooks, A.~Van Den~Hengel, and D.~Gawley, ``On the fitting of surfaces to data with covariances,'' \emph{IEEE Transactions on pattern analysis and machine intelligence}, vol.~22, no.~11, pp. 1294--1303, 2000.

\bibitem{barber1996quickhull}
C.~B. Barber, D.~P. Dobkin, and H.~Huhdanpaa, ``The quickhull algorithm for convex hulls,'' \emph{ACM Transactions on Mathematical Software (TOMS)}, vol.~22, no.~4, pp. 469--483, 1996.

\bibitem{saalfeld1999topologically}
A.~Saalfeld, ``Topologically consistent line simplification with the douglas-peucker algorithm,'' \emph{Cartography and Geographic Information Science}, vol.~26, no.~1, pp. 7--18, 1999.

\bibitem{mur2017orb}
R.~Mur-Artal and J.~D. Tard{\'o}s, ``Orb-slam2: An open-source slam system for monocular, stereo, and rgb-d cameras,'' \emph{IEEE transactions on robotics}, vol.~33, no.~5, pp. 1255--1262, 2017.

\bibitem{Jocher_Ultralytics_YOLO_2023}
\BIBentryALTinterwordspacing
G.~Jocher, A.~Chaurasia, and J.~Qiu, ``{Ultralytics YOLO},'' Jan. 2023. [Online]. Available: \url{https://github.com/ultralytics/ultralytics}
\BIBentrySTDinterwordspacing

\bibitem{kirillov2023segment}
A.~Kirillov, E.~Mintun, N.~Ravi, H.~Mao, C.~Rolland, L.~Gustafson, T.~Xiao, S.~Whitehead, A.~C. Berg, W.-Y. Lo, \emph{et~al.}, ``Segment anything,'' in \emph{Proceedings of the IEEE/CVF International Conference on Computer Vision}, 2023, pp. 4015--4026.

\bibitem{song2022scale}
S.~Song, J.~Zhao, T.~Feng, C.~Ye, and L.~Xiong, ``Scale estimation with dual quadrics for monocular object slam,'' in \emph{2022 IEEE/RSJ International Conference on Intelligent Robots and Systems (IROS)}.\hskip 1em plus 0.5em minus 0.4em\relax IEEE, 2022, pp. 1374--1381.

\bibitem{sturm2012benchmark}
J.~Sturm, N.~Engelhard, F.~Endres, W.~Burgard, and D.~Cremers, ``A benchmark for the evaluation of rgb-d slam systems,'' in \emph{2012 IEEE/RSJ international conference on intelligent robots and systems}.\hskip 1em plus 0.5em minus 0.4em\relax IEEE, 2012, pp. 573--580.

\bibitem{handa2014benchmark}
A.~Handa, T.~Whelan, J.~McDonald, and A.~J. Davison, ``A benchmark for rgb-d visual odometry, 3d reconstruction and slam,'' in \emph{2014 IEEE international conference on Robotics and automation (ICRA)}.\hskip 1em plus 0.5em minus 0.4em\relax IEEE, 2014, pp. 1524--1531.

\bibitem{de2000computational}
M.~De~Berg, \emph{Computational geometry: algorithms and applications}.\hskip 1em plus 0.5em minus 0.4em\relax Springer Science \& Business Media, 2000.

\bibitem{lee2017shoelace}
Y.~Lee and W.~Lim, ``Shoelace formula: Connecting the area of a polygon and the vector cross product,'' \emph{The Mathematics Teacher}, vol. 110, no.~8, pp. 631--636, 2017.

\end{thebibliography}
\end{document}